\documentclass[12pt]{article}
\usepackage{amsmath}
\usepackage{amsfonts}
\usepackage{amssymb}
\usepackage{amsthm}
\usepackage{setspace}
\usepackage{xcolor}
\usepackage{graphicx}
\usepackage{hyperref}
\usepackage{authblk}
\usepackage[numbers, sort, comma, square]{natbib}

\providecommand{\keywords}[1]
{
  \small	
  \textbf{\textit{Keywords---}} #1
}

\title{Vision-based Estimation of Fatigue and Engagement in Cognitive Training Sessions}
\author[1]{Yanchen Wang\footnote{Equal contribution first co-authors}}
\author[1]{Adam Turnbull$^*$}
\author[3]{Yunlong Xu}
\author[4]{Kathi Heffner}
\author[1]{Feng Vankee Lin}
\author[1,2]{Ehsan Adeli\thanks{Email address: \texttt{eadeli@stanford.edu}; Corresponding author}}
\affil[1]{Department of Psychiatry and Behavioral Sciences, 
      Stanford University, 
      Stanford, California, USA}
\affil[2]{Department of Computer Science, 
      Stanford University, 
      Stanford, California, USA}
\affil[3]{Department of Neurobiology, 
      University of Chicago, 
      Chicago, Illinois, USA}
\affil[4]{School of Nursing, 
      University of Rochester, 
      Rochester, New York, USA}

\date{}

\begin{document}
\maketitle
\begin{abstract}
Computerized cognitive training (CCT) is a scalable, well-tolerated intervention that has promise for slowing cognitive decline. Outcomes from CCT are limited by a lack of effective engagement, which is decreased by factors such as mental fatigue, particularly in older adults at risk for dementia. There is a need for scalable, automated measures that can monitor mental fatigue during CCT. Here, we develop and validate a novel Recurrent Video Transformer (RVT) method for monitoring real-time mental fatigue in older adults with mild cognitive impairment from video-recorded facial gestures during CCT. The RVT model achieved the highest balanced accuracy($78\%$) and precision (0.82) compared to the prior state-of-the-art models for binary and multi-class classification of mental fatigue and was additionally validated via significant association ($p=0.023$) with CCT reaction time. By leveraging dynamic temporal information, the RVT model demonstrates the potential to accurately measure real-time mental fatigue, laying the foundation for future personalized CCT that increase effective engagement.
\end{abstract}


\section{Introduction} 

Computerized cognitive training (CCT) is an essential front-line intervention for preventing or slowing cognitive decline and brain aging seen in dementia~\citep{national2017preventing} and has great promise given that it is scalable, personalizable, and well-tolerated. However, research findings on the effectiveness of CCT vary, and although there is significant variability in adherence~\citep{turunen2019computer} (measured via time spent on training), there is also no clear dose-response relationship in CCT: some studies have even found that increasing training time may lead to worse outcomes~\citep{lampit2014computerized}. This complicated finding suggests that time spent \textit{effectively} engaging~\citep{yardley2016understanding} in CCT is more important than adherence per se~\citep{stern2012cognitive}. Designing more effective CCT interventions requires an understanding of the balance between increasing training time and maintaining effective engagement. 

Particularly in older adults at risk for dementia, too much training may lead to increased mental fatigue, that is, the acute decline in mental status, including attention, motivation, energy, and alertness, that occurs during tasks demanding prolonged effort~\cite{lin2022multi}. Mental fatigue interferes with cognitive plasticity and training gains, as mental fatiguability is increased in those at-risk for dementia, including those with mild cognitive impairment; MCI~\citep{kukla2022brain}). Mental fatigue is difficult to measure using self-report, which requires regular task interruption to collect. Objective measures that relate to mental fatigue (e.g., CCT task performance) are affected by many factors other than mental fatigue. Developing automated tools to capture mental fatigue during CCT is therefore critical to monitor and improve effective engagement. Additionally, automated measures of mental fatigue are important for developing scalable and accessible CCT that can be administered at home to older adults at risk for dementia without the need for family member or professional interventionist supervision. In addition to the lack of availability of these resources, research suggests that at-home CCT is less effective~\citep{lampit2014computerized}, potentially due to decreased effective engagement that limits the potential of CCT in certain groups (e.g., those living alone) that are at higher risk for dementia~\citep{huang2023social}, contributing to health disparities~\citep{hill2015national} in dementia risk and outcomes. Monitoring and improving effective engagement using automated tools that can be applied to at home CCT is therefore of high clinical and public health relevance.

\begin{figure}[t]
    \centering
    \includegraphics[width=0.98\linewidth,trim={5 17 5 5},clip]{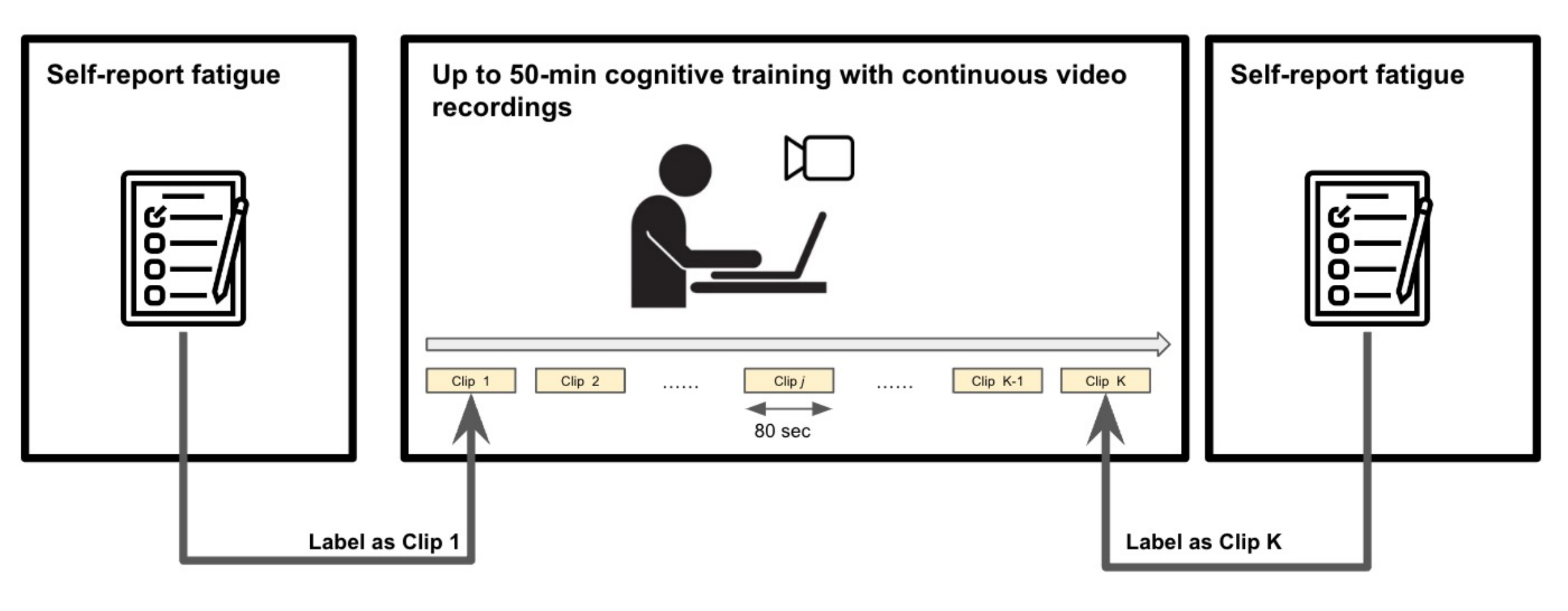} \vspace{-12pt}
    \caption{Experimental flow for a weekly in-person CCT session. As part of a CCT clinical trial conducted during COVID-19, each participant completed weekly in-person training sessions (where performance, via reaction time in tasks, were recorded) throughout an 8-week CCT. Self-report fatigue (an 18-item visual analogue scale) was assessed at the beginning (pre-fatigue) and end (post-fatigue) of the training session with video recording continuously monitoring facial gestures. To train our model, we used pre- and post-fatigue measures as the ground-truth labels for the first and the last clips.   
    }
    \label{fig:training}
\end{figure}

In this paper, we propose a method for predicting real-time mental fatigue from video-recorded facial expressions during CCT in older adults with MCI. This scalable method monitors mental fatigue during CCT and can be easily adapted in future closed-loop CCTs to manage mental fatigue and increase effective engagement. Previous work suggests that mental fatigue disrupts neutral facial expression~\citep{mlynski2021fatigue,kong2021facial}, and that selected facial expression patterns reflect mental fatigue induced by tasks requiring sustained attention~\citep{kong2021facial,gu2002active,li2019accurate}. We therefore propose to capture patterns of facial expression reflecting real-time mental fatigue as a way of quantifying and improving effective engagement. For a CCT session, outlined in Figure~\ref{fig:training}, we video-record the individual's facial gestures throughout the session. Self-report pre- and post-session fatigue were assessed and the participants' performance was recorded through reaction time (RT) in tasks. 

\begin{figure}[t]
    \centering
    \includegraphics[width=\linewidth]{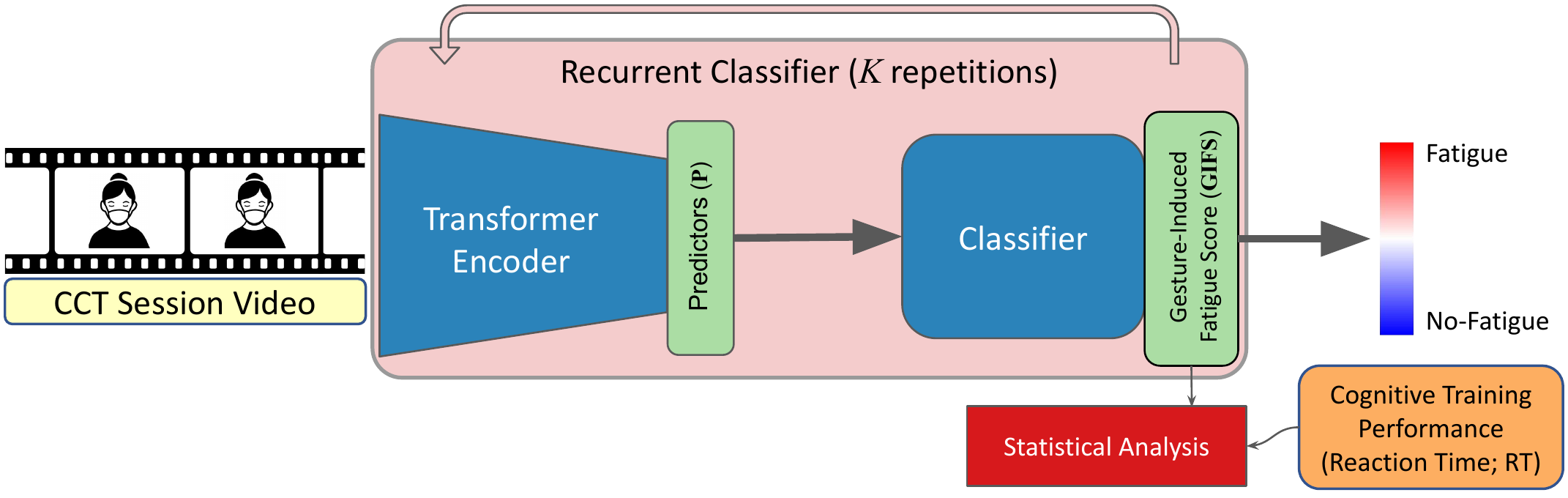}\vspace{-10pt}
    \caption{Analytical flow. Our novel transformer-based computer vision model extracts predictors (P) and then outputs a Gesture-Induced Fatigue prediction Score (or GIFS for short), which in turn is classified into Fatigue or No-Fatigue. After training the model in an end-to-end fashion, this setting allows for rigorous statistical analysis between GIFS and the CCT performance index (reaction time; RT), through which we validate the relationship between facial expression and CCT task performance.}
    \label{fig:framework}
\end{figure}

Using this data, we used recent advances in computer vision and developed a novel Recurrent Video Transformer (RVT), combining a clip-wise transformer encoder module \cite{dosovitskiy2021an} and a session-wise Recurrent Neural Network (RNN) \cite{sherstinsky2020fundamentals} classifier. Specifically, the model involved a transformer-based encoder that reduces each video clip to a series of predictors (denoted by \textbf{P}) and a classifier to turn them into a prediction score, which we refer to as Gesture-Induced Fatigue Scores (\textbf{GIFS}). As depicted in Figure~\ref{fig:framework}, this process is repeated for each of the \textit{K} sub-clips of the entire session video, hence the recurrence. We quantified performance of RVT via accuracy, $F_1$ score, and precision for predicting a well-validated measure of mental fatigue. Our model outperformed previous state-of-the-art models for both detecting mental fatigue (binary classification) and rating the level of mental fatigue (multi-class classification). Finally, we used an objective measure of performance (i.e., RT) that is known to increase with increasing mental fatigue~\citep{jaydari2019mental} to validate the facial expression-derived fatigue scores. 

\section{Methods}
\label{sec:methods}

\subsection{Participants}
\label{sec:study}


\begin{table}[t]
\centering
\caption{Mean and standard deviation (Stdev) study participants' demographic information, MoCA scores, and session length (in minutes).}
\label{tab:demo}
\begin{tabular}{lcccc}
\hline
\textbf{Variable}   & \textbf{Total} & \textbf{Mean} & \textbf{Stdev} \\ \hline
N & 24 &  &  \\
Age                &        & 72.67        & 8.34           \\
Years of education   &   & 15.46        & 2.48        \\
Sex (F/M) & 16/8 & - & - \\
\hline
MoCA & 24         & 23.79        & 2.32        \\
Session length (min) & 111           & 57.37 & 8.47 \\ \hline
\end{tabular}
\end{table}

In an ongoing clinical trial of computerized cognitive training in older adults with MCI \citep{lin2021targeting}, 
we video-recorded participants’ facial expressions throughout each in-lab 50-minute training session weekly with up to 8 sessions per participant. Participants reported their perceived mental fatigue with an 18-item visual analog scale at the beginning (pre-fatigue) and end (post-fatigue) of the session. Each training section also includes 5 different tasks. Each task contains multiple trials. We record each trial RT in the training.

The study was approved IRB by the University of Rochester and Stanford University. As shown in Table~\ref{tab:demo}, The dataset contains data from 24 participants who had at least one complete 50-minute training session video data (mean age of 72.67 years, 15 females/9 males, mean years of education at 15). Participants' global cognition was assessed at baseline using Montreal Cognitive Assessment (MoCA)  \citep{nasreddine2005montreal}, with a mean score of 23.79. There were a total of 111 sessions of video data across all individuals. On average, each participant contributed 5 sessions, and each session's duration was 57.4 minutes.

In addition to the above clinical dataset, we use a public large-scale facial expression dataset, Dynamic Facial Expressions in the Wild (DFEW)~\citep{10.1145/3394171.3413620}. This dataset is used to pre-train our encoder that extracts predictors \textbf{P}. 


\subsection{Measure of Mental Fatigue}
Mental fatigue was measured using an 18-item visual analogue scale, which is the most widely used fatigue state measure in the literature, 
including in multiple studies of fatigue and cognitive impairment~\citep{lee1991validity,lin2016mental}. The scale evaluates various aspects of disengagement in domains of affect, motivation, and attention; the participants indicate their responses using a 0-10 scale. An average score across all items was calculated with higher values indicating more fatigue. Fatigue data were collected at two-time points within a training session: the beginning (pre-fatigue) and the end (post-fatigue) of a session.

\begin{figure}
    \centering
    \includegraphics[width=1\linewidth,trim={10 60 20 60},clip]{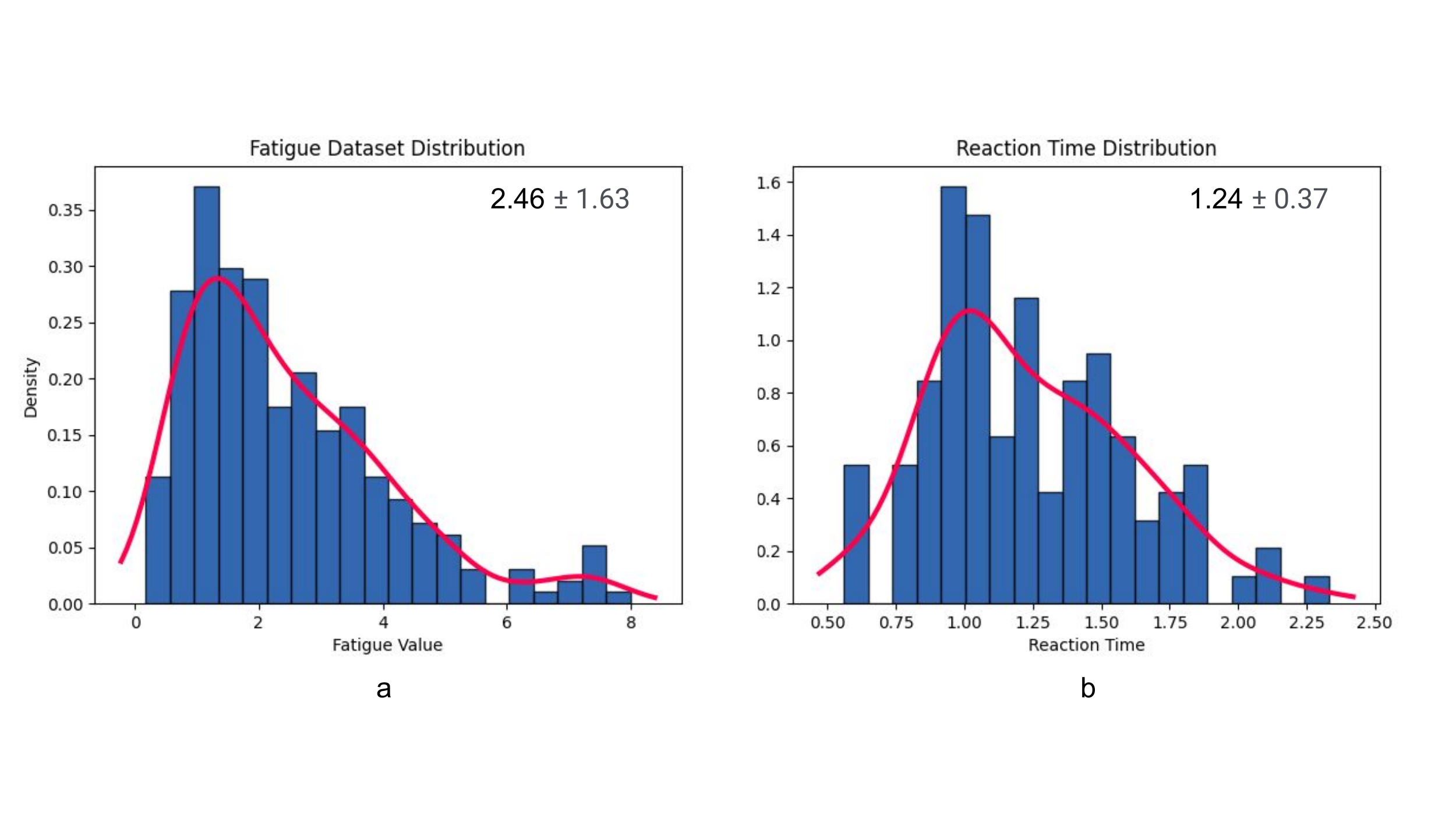}\vspace{-10pt}
    \caption{(a) Distribution of the Fatigue value, merging pre- and post-fatigue scores from all sessions and all participants in our dataset, with the x-axis representing the assigned fatigue values for each video, and the y-axis indicating the density of each fatigue value's occurrence in the dataset. (b) Distribution of Reaction Time. The curve represents the kernel density estimate (KDE) used to fit the data distribution.}
    \label{fig:distribution}
\end{figure}

Fig.~\ref{fig:distribution}(a) shows the distribution of ground-truth Fatigue values, merging pre- and post-fatigue scores, from all sessions and all participants in our dataset. Additionally, we recorded participants' CCT performance index in terms of reaction time (RT), summarized in Fig.~\ref{fig:distribution}(b). As it is also visible in the plots, these two metrics are strongly correlated (\textit{p}-value$=0.007$ in Section~\ref{sec:rt}). This observation supports the hypothesis that RT records the individual's performance and is highly associated with mental fatigue. 

Our primary objective is to predict the fatigue level based on video-recorded facial gestures. However, treating the entire video as input is inefficient and redundant as each original video is 50 minutes long with a frame rate of 30 frames per second (FPS). Moreover, the participant's pose with respect to the camera is quite stable throughout the video, and too much of the same and similar data could cause a loss of plasticity in the model training \cite{masel2007loss}. Therefore, we downsampled the videos to one frame every five seconds. 


\subsection{Recurrent Vision Transformer (RVT)} 

\begin{figure}[t]
    \centering
    \includegraphics[width=1\linewidth,trim={0 0 0 0},clip]{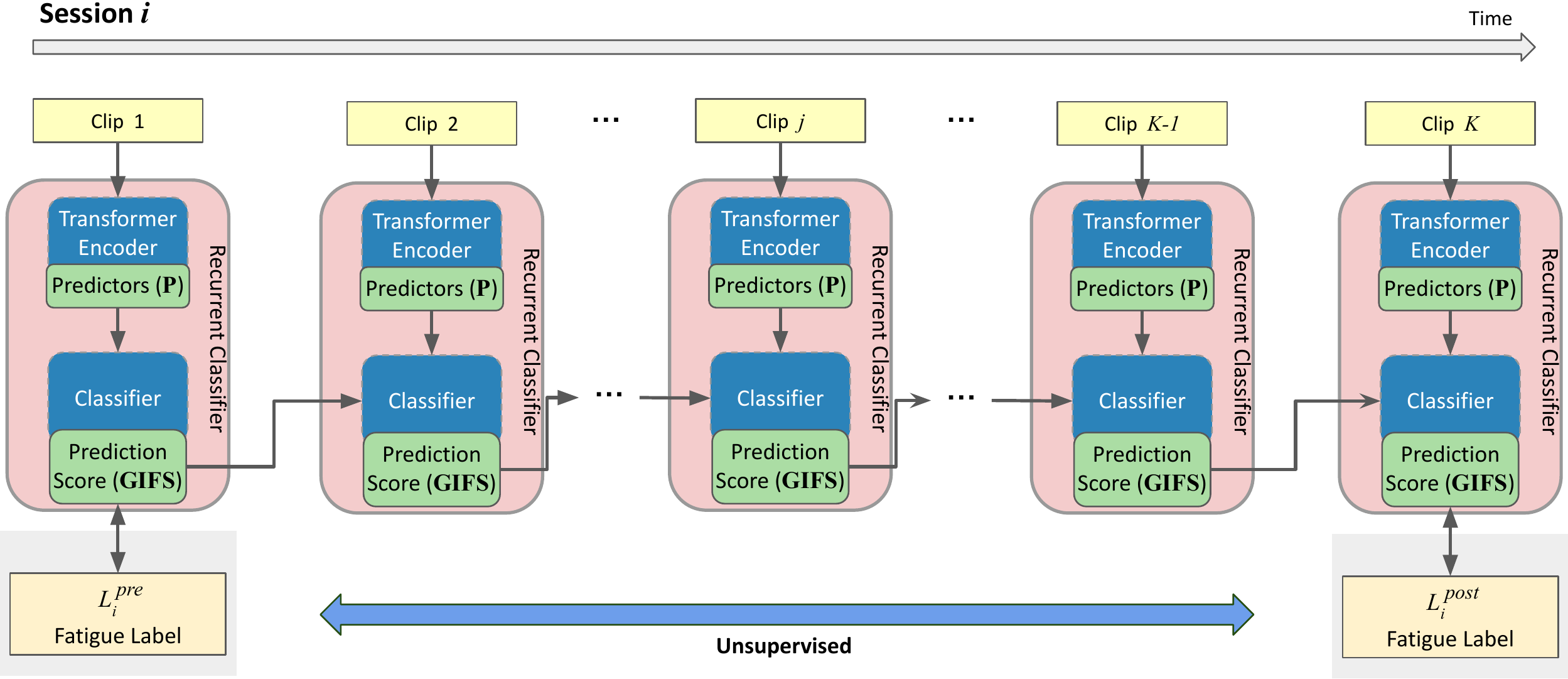} \vspace{-15pt}
    \caption{Our proposed Recurrent Vision Transformer (RVT) for fatigue detection. We used this model for both binary (detect the presence of fatigue) and multi-class (estimate the degree of fatigue) classification. }
    \label{fig:rvt}
\end{figure}

As shown in Fig.~\ref{fig:rvt}, our proposed RVT model consists of a clip-wise transformer encoder module which includes a convolutional spatial transformer, a temporal transformer, and a session-wise Recurrent Neural Network (RRN) classifier. The high-level idea is that for each session, the clips within each session are encoded by the transformer encoder which is detailed in Fig~\ref{fig:tvc}. It then predicts the fatigue label via a classifier in the RNN cell. Note that because of the study design, we only have ground-truth Fatigue labels for the first and last clips, defining our only sources of supervision. We do not compute the loss for the clips in between but pass the recurrent activity. 

\noindent\textbf{Input.} 
Each input sample is a clip of $X \in \mathcal{R}^{T \times 3 \times H \times W }$, where $T$ indicates the number of frames, $H$ the height of each frame, and $W$ the width of each frame. Specifically, within each session $O_i \in \mathcal{R}^{E \times 3 \times H \times W}$, there are $K$ clips, namely $C_i=\{X_{i1}, X_{i2}, \cdots, X_{iK}\}$. Suppose we have $\alpha$ training sessions in total in the dataset, i.e., $i \in \{1,\ldots, \alpha\}$. For any $O_i$, there are two corresponding labels $L_{i}^{pre}\in R$ and $L_{i}^{post} \in R$ indicates the Fatigue value immediately before the training session and the Fatigue value immediately after a training session, respectively. We assign the $L_{i}^{pre}$ as the label of $O_i^{pre}$ and $L_{i}^{post}$  as the label of $O_i^{post}$. For any $X_{ij}$, where $2 \leq j \leq K-1, i \in \mathcal{N}$, we do not use any ground-truth labels. We hence divide our dataset into two parts $V=\{X\mid X=O_{ij} \in C_i , j = 1 \, or \, K, \forall i\}$ (labelled parts), and $U=\{X\mid X=O_{ij} \in C_i , 2 \leq j \leq K-1, \forall i\}$ (unlabelled parts). We use $\mathcal{X}$ to denote $V \cup U$. Note that the length of each session is roughly 50 minutes (a long $\sim$50-min video). As known in the computer vision community, transformers require large datasets for training; hence, to enable training the models on our small clinical dataset, we sub-sample each session-long video into several non-overlapping sub-clips, augmenting the training process. These sub-clips are then connected with each other using the RNN. Here, we subsample each session into $80$-second clips ($2400$ frames); we uniformly sample 1 frame for every 5 seconds and obtain exactly $16$ frames in each $X$, which means $T=16$. 

\noindent\textbf{Transformer Encoder}
As shown in Figure~\ref{fig:tvc}, the described system involves a transformer encoder component made up of three main modules: a face detection module equipped with a face mask detector, a spatial transformer encoder, and a temporal transformer encoder. Initially, the face detection module, based on ResNet50~\citep{he2016deep}, identifies and crops the face in an image and detects if a face mask is present. This module produces the input of the transformer module. Secondly, the spatial and temporal transformer encoders, pre-trained on specific datasets~\citep{10.1145/3394171.3413620} by \cite{zhao2021former}, to recognize seven basic emotions from video clips. The spatial transformer takes processed face information, extracts features through convolution blocks, and then, using a self-attention mechanism, feeds into the temporal encoder, resulting in encoded features representing the facial expressions in a given video clip (details in the Appendix). The encoder summarizes the sequence input facial images into a predictors vector \textbf{P}.


\noindent\textbf{Classification.}
As depicted in Figure~\ref{fig:rvt} and \ref{fig:tvc}, for each video clip $X \in \mathcal{X}$, the RNN module serves as a classifier, taking the output $X^E$ from the Transformer encoder and predicting a fatigue value $f(x) \in {0,1}$ (binary output). To perform this prediction, the classifier first turns \textbf{P} into a Gesture-Induced Fatigue Score (\textbf{GIFS}) and then binerizes it to $f(x)$. During training, video clips are fed into the network session by session. Within each session, clips are input sequentially, following their original order in the video. This approach enables the network to track the dynamics of fatigue within a session and accumulate long-term temporal information through recurrent connections. Specifically, in each epoch, we feed the network with $O_1, O_2, \cdots, O_i, O_{i+1},\cdots, O_{\alpha}$, in increasing order of $i$. For any given $O_i$, we compute the loss of $X_{ij}$ based on $O_i^{pre}$ and $O_i^{post}$, if $j=1$ or $K$, respectively. For any $X_{ij}$, where $1<j <K$, we do not compute the loss and perform unsupervised training. Hence, the overall loss function $\mathcal{L}$ for training the model is defined as:
\begin{equation} 
    \mathcal{L} = \sum_{i=1}^{\alpha} (\mid f(X_{i1})-O_i^{pre} \mid + \mid f(X_{iK})-O_i^{post} \mid )
    \label{eq:loss_func}
\end{equation}

Note that for $X_{ij}$, if $1<j <K$, the RNN cells receive the recurrent activities corresponding to $X_{i,j-1}$ and pass their recurrent activities to the RNN classifier corresponding to $X_{i,j+1}$. If $j=1$, the RNN does not receive any recurrent activities and only passes its recurrent activities to the next time point. Similarly, if $j=K$, the RNN does not pass any recurrent activities and only receives the recurrent activities from the previous time point.

\noindent\textbf{Evaluation.}
To create our training set, we extracted 16-frame clips by segmenting the video sessions as described earlier. There are two different parts of video clips, labeled parts $V$ and unlabelled parts $U$. In total, there are $860$ clips in $\mathcal{X}=U \cup V$. 



We used a participant-level Leave-one-out-cross-validation (LOOCV) technique for training and evaluation. Specifically, our LOOCV setting leaves one entire participant out (i.e., videos of all sessions from that participant), trains the model on all other participants, and tests the all clips of the left-out participant. This is repeated for all participants and mean±standard-deviation of all evaluation metrics are reported. This procedure allows for assessing the model generalization performance on new and unseen subjects while ensuring no bias by observing the same subject in a different training session. 


To ensure the reproducibility of our experiments and fair comparison between models, we implemented random seeding. Specifically, we ensured that all our experiments were conducted under the same seed conditions to enable a like-for-like comparison across the competing methods. 
By using the same seed values across all experiments, we ensured that the random initialization of our models was consistent. 
Additionally, the learning rate is set to $5e^{-6}$, the clip batch size to $1$, and the number of epochs to $40$ during training. 


\noindent\textbf{Validation Against Reaction Time.} 
During individual training sessions, RTs to accurate responses to cognitive training tasks were averaged within a task. We then averaged these task-based mean RTs to generate a session-based RT for use in validation analyses by associating them with session-based facial expression fatigue composite scores (\textbf{GIFS} in Figure~\ref{fig:framework}). Since individual tasks require different mean RTs (e.g., due to the requirement to make multiple vs. single selections), we only included session-based RTs from those sessions with completed 5 training tasks to ensure the comparability between RTs across participants.

To examine the validity of facial expression fatigue composite scores (\textbf{GIFS}), we assessed the known relationship between increased fatigue (measured via \textbf{GIFS}) and worse performance (measured via increased RT). 
We conducted Generalized Estimating Equation analyses using the following equations employing AR(1) correlational matrix taking session-based reaction (RT) as outcomes:
\begin{align}
    \text{RT}_{i}&=i +\epsilon, 
    \label{eq:rt1}\\
    \text{RT}_{i}&=i+L^{mean}_{i}+\epsilon, 
    \label{eq:rt2}\\ \text{RT}_{i}&=i+L^{mean}_{i}+\text{GIFS}_{i}+\epsilon, 
    \label{eq:rt3}
\end{align}
where $i$ refers to the session index and $\epsilon$ indicates error. Eq.~\eqref{eq:rt1} tested whether RT improved over sessions as seen in the literature~\citep{chen2020autonomic}; Eq.~\eqref{eq:rt2} was to examine whether self-report fatigue related to RT across sessions; and Eq.~\eqref{eq:rt3} was to evaluate the relationship between the facial expression composite fatigue score (GIFS) and RT across sessions considering the effect of the number of sessions and self-report fatigue: $L_{mean}$. Of note, $L_{mean}$ here refers to the average of fatigue between the beginning and end of a session.

\section{Results}

We evaluate the performance of our model on our dataset in a leave-one-out-cross-validation setting using their balanced accuracy (to account for possible class imbalance in multi-class settings), $F_1$ score, precision, and area under ROC curve (AUC). 
To illustrate the superiority of the proposed method compared to the prior work, we compare it with 
Video Transformer (VT) \citep{zhao2021former}, Video Transformer with sequential data input (VT-Seq), ResNet-34, ResNet-34 with sequential data input (ResNet-Seq) \citep{he2016deep}, EM-CNN (Eye-Mouth CNN) \citep{zhao2020driver}, and PFLD (Practical Facial Landmark Detector) models \citep{guo2019pfld}. 
Note that, for the sake of this comparison, we fine-tuned a ResNet-34 model on our data, which was pre-trained on FER2013 \citep{he2016deep,pham2021facial, goodfellow2013challenges}. We additionally trained the EM-CNN and PFLD models on our fatigue dataset.


\subsection{Binary Classification}

To conduct binary Fatigue classification, we split the Fatigue scores into two classes. The median of the Fatigue score in our dataset was 2 (see Figure~\ref{fig:distribution}). To define the classes, we use 2 as the cutoff threshold, i.e., Fatigue score $< 2$ defines class 0, and Fatigue score $\geq 2$ defines class 1. This setting creates a classification problem with 117 samples for class 0 and 131 samples for class 1. The results of our binary classification experiment are presented in Table~\ref{tab:fatigue_results}. Our approach obtains a significantly better performance compared to others in terms of average accuracy, $F_1$ score, and precision. This significant gain in performance could be due to the fact that our method intelligently uses the middle clips (those without ground-truth labels) to train the model in a recurrent way, implicitly augmenting the data, while the supervision signal only comes from pre- and post-fatigue labels. 

\begin{table}[t]
\centering
\caption{Comparison of our RVT results with the previous state-of-the-art and several other methods implemented and validated by us, tested on our Fatigue dataset. The best results are typeset in bold. * marks methods that are significantly worse than our method ($p<0.05$) measured by the Wilcoxon signed rank test \citep{wilcoxon1992individual}.}
\vspace{5pt}
\begin{tabular}{lcccc}
\hline
\textbf{Model} & \textbf{Accuracy (\%)} & \textbf{F1 Score} & \textbf{Precision} & \textbf{AUC}\\ \hline
RVT (Ours) & \textbf{79.58 ± 3.42} & \textbf{0.79 ± 0.03} & \textbf{0.82 ± 0.03} & \textbf{0.80 ± 0.03} \\
VT$^\ast$ & 66.37 ± 3.21 & 0.68 ± 0.04 & 0.78 ± 0.04 & 0.66 ± 0.03\\
VT-Seq$^\ast$ \citep{zhao2021former} & 68.92 ± 2.87 & 0.70 ± 0.04 & 0.81 ± 0.05 & 0.69 ± 0.03\\
ResNet-34$^\ast$ & 70.27 ± 0.97 & 0.71 ± 0.01 & 0.78 ± 0.003 & 0.70 ± 0.01\\
ResNet-Seq$^\ast$ & 75.08 $\pm$ 2.25  & 0.75 $\pm$ 0.03 & 0.81 $\pm$ 0.03 & 0.75 $\pm$ 0.03 \\
EM-CNN$^\ast$ \citep{zhao2020driver} & 71.32 ± 1.06 & 0.69 ± 0.01 & 0.76 ± 0.01 & 0.71 ± 0.01\\ 
PFLD$^\ast$ \citep{guo2019pfld} & 69.52 ± 4.31 & 0.68 ± 0.06 & 0.77 ± 0.02 & 0.70 ± 0.07\\ \hline
\end{tabular}
\label{tab:fatigue_results}
\end{table}

For the models that took images as input, such as ResNet-34, EM-CNN, and PFLD \citep{he2016deep, zhao2020driver, guo2019pfld}, we set the batch size equal to the clip frame size, which is $K = 16$ in our case. We use the same label for all frames in the clip, which is equal to the ground-truth label of the clip. Since only the first and the last clips are labeled with fatigue scores, they are used to train the non-sequential competing methods.  
Note that the suffix ``Seq" in the table indicates that the model input was sequential, with the video clips sorted in time series order.


\subsection{Explainability}
\begin{figure}[t]
    \centering
    \includegraphics[width=\linewidth,trim={10 45 10 40},clip]{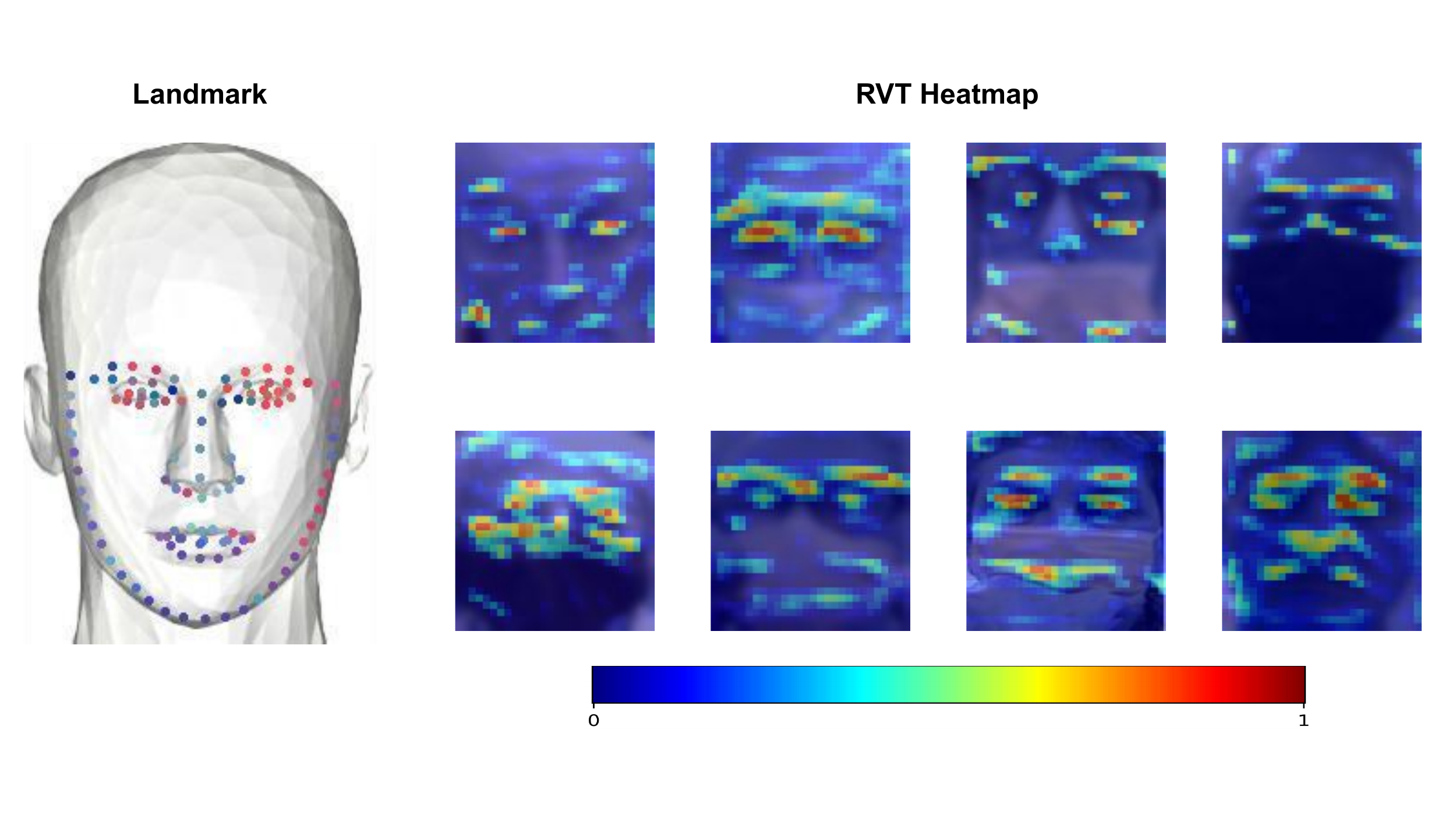}\vspace{-10pt}
    \caption{Facial keypoints captured by our RVT model reflecting fatigue and disengagement. Left: a template face with colored landmarks representing the average value of the corresponding heatmap regions across the entire dataset; Right: heatmaps from selected participants. The color indicates the contribution of those pixels/regions in correctly predicting mental fatigue.}
    \label{fig:heatmap}
\end{figure}

The primary contribution of this paper is the introduction of an encoder-based model (Figures~\ref{fig:rvt} and \ref{fig:tvc}) for estimating fatigue in long videos with only a small amount of ground-truth pre- and post-fatigue labels. But to gain insights into the model and understand to which parts of the image it attended to correctly estimate fatigue, we visualize the model saliencies for each individual. Specifically, we create a heatmap on the participant's face highlighting which pixels the model has attended to for estimating fatigue correctly. The results of sample participants are visualized in Figure~\ref{fig:heatmap}(right). Additionally, Figure~\ref{fig:heatmap}(left) shows a template face with predefined landmarks. To generate this figure, the face landmarks are detected for all our patients in all frames, using FaceX-Zoo~\citep{wang2021facex}. The attention levels around those points in the heatmaps generated by our RVT are then averaged. Hence, each landmark point is colored corresponding to its contribution significance to correctly detecting fatigue over our entire dataset. The contribution significance values are normalized to the range [0, 1] (following the colormap) for visualization.  

With a primary focus on areas such as the eyes and mouth, RVT's attention mechanism significantly contributes to its superior performance in fatigue classification when compared to state-of-the-art methods. As evidenced by Figure~\ref{fig:heatmap}~(left), more attention is given to regions around the eye, than to the mouth areas, since a large number of the study participants were masked during their cognitive training sessions. These results validate that our method interpretably understands fatigue and hence could further be used to inform user engagement in the training.  

\subsection{Estimating the Degree of Fatigue (Multi-Class Classification)} 

The label values in the Fatigue score in our dataset were found to be skewed, as shown in Figure~\ref{fig:distribution}(a), and therefore a cutoff of 2 was selected in the binary classification case. In addition to the Binary Classification approach presented in Table~\ref{tab:fatigue_results}, we implemented a 3-Class Classification approach to rate the degree of Fatigue in individuals. Similar to the 2-Class Classification, we defined the class label $\mathcal{C}$ according to the threshold defined here:
\begin{equation}
     \mathcal{C} =  \left\{\begin{array}{l}
     0 \text{ if } v < 2,\\
     1 \text{ if } 2 \leq v < 5,\\
     2 \text{ otherwise},
    \end{array}\right.
    \label{eq:3-class-label-value}
\end{equation}
where $v$ is the value of the self-report Fatigue value. This setting creates three classes with 106 samples in class 0, 100 in class 1, and 16 in class 2, introducing a high imbalance for classes 1 and 2 compared to 0.
\begin{figure}[t]
    \centering
    \includegraphics[width=0.5\linewidth]{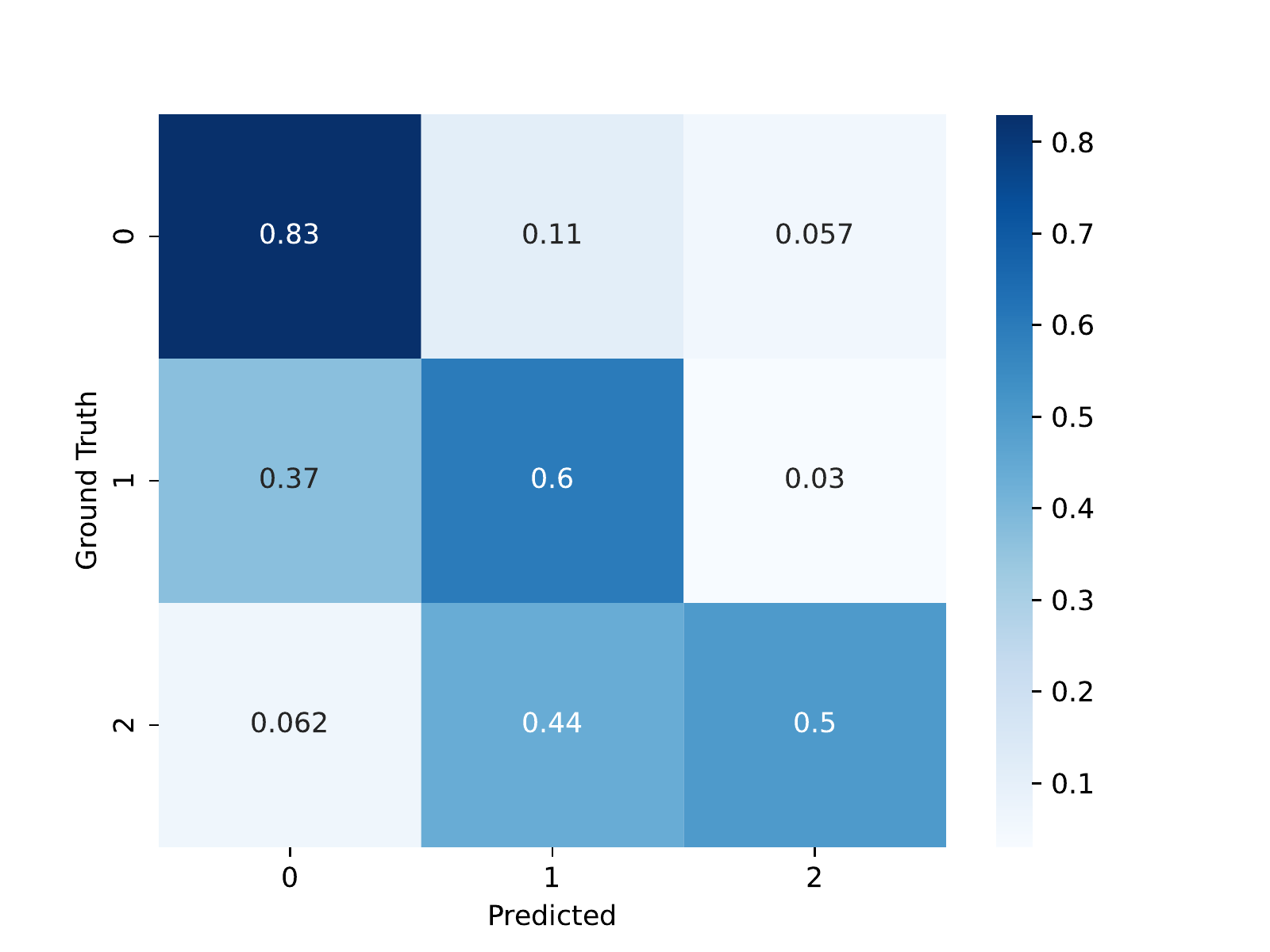}\vspace{-10pt}
    \caption{Confusion Matrix of our RVT model. }
    \label{fig:confusion_matrix}
\end{figure}

The results and comparisons are summarized in Table~\ref{tab:fatigue_results_3}. Our proposed model RVT successfully achieves a balanced accuracy of 59.78\%, which significantly outperforms the previous state-of-the-art models, when they are run on our dataset in a multi-class setting. As indexed by the balanced accuracy (BAcc), other methods suffer from the class imbalance induced by classes 1 and 2, while our method shows less sensitivity. The confusion matrix of multi-class RVT displaying accuracy is shown in Figure~\ref{fig:confusion_matrix}. As shown, RVT can more easily identify class 0 (not fatigue) compared to the other two classes. This may be explained by the smaller number of samples in the other two classes and higher levels of imbalance. 

\begin{table}\centering \footnotesize
\caption{Results of three classification methods. In evaluating these methods, we use balanced accuracy (BAcc) to account for our imbalanced data, in addition to $F_1$ score, precision, and the area under the curve (AUC). Best results are typeset in bold and $\ast$ indicates methods that are significantly worse than ours ($p \leq 0.05$) measured by the Wilcoxon signed rank test \citep{wilcoxon1992individual}.}
\vspace{5pt}
\begin{tabular}{lcccc}
\hline
\textbf{Model} & \textbf{BAcc (\%)} & \textbf{F1 Score} & \textbf{Precision} & \textbf{AUC} \\ \hline
RVT (Ours) & \textbf{59.78 $\pm$ 2.38} & \textbf{0.68 $\pm$ 0.02} & \textbf{0.74 $\pm$ 0.05} & \textbf{0.73 $\pm$ 0.004} \\  
EM-CNN$^\ast$ \citep{zhao2020driver} & 55.65 $\pm$ 2.08 & 0.64 $\pm$ 0.02 & 0.67 $\pm$ 0.01 & 0.70 $\pm$ 0.02 \\ 
PFLD$^\ast$ \citep{guo2019pfld} & 55.35 $\pm$ 1.97 & 0.55 $\pm$ 0.03 & 0.68 $\pm$ 0.02 & 0.61 $\pm$ 0.04 \\ \hline 
\end{tabular}
\label{tab:fatigue_results_3}
\end{table}

\subsection{CCT Task Performance/Reaction Time}
\label{sec:rt}
RT is measured as the mean duration it takes for a subject to respond on trials where they made accurate responses during a session. To examine the validity of facial expression fatigue composite scores (i.e., GIFS), we performed several Generalized Estimating Equations with AR(1) correlational matrix analyses assessing well-established relationships using our session-based data (fatigue from self-report and facial expressions, and performance via RT). We included  85 complete sessions from 23 participants for these analyses. As expected, there was a significant improvement across sessions in terms of reduction of RT ($B=-0.03, SE=0.01, \text{Wald }\chi^{2}=9.06, p=0.003$) (Eq.~\eqref{eq:rt1} in Section~\ref{sec:methods}). This result helps to validate our measure of RT as reflecting performance. Lower self-report fatigue was significantly related to lower RT (better performance) over sessions ($B=0.024, SE=0.01, \text{Wald }\chi^{2}=7.17, p=0.007$) (Eq.~\eqref{eq:rt2} in  Section~\ref{sec:methods}). Lower facial expression fatigue composite score (GIFS) was related to better performance, as lower RT, over sessions ($B=0.03, SE=0.01, \text{Wald }\chi^{2}=5.16, p=0.023$) (Eq.~\eqref{eq:rt3} in Section~\ref{sec:methods}), taking into account session number and self-reported fatigue. These findings validate our measures of fatigue and performance and suggest that facial expression fatigue captures aspects of fatigue important for performance over and above those captured by self-report.

\section{Discussion}

\paragraph{Effective Engagement in Cognitive Training.} 
In this paper, we developed a method for automated monitoring of real-time mental fatigue via participant facial expressions during CCT in older adults with MCI. Mitigating or preventing mental fatigue is critical to increasing effective engagement in CCT, particularly in older adults at risk for dementia. By developing scalable, real-time measures of mental fatigue, we provided a foundation for future closed-loop CCT that can adapt to manage mental fatigue, allowing for increased time spent on training while maintaining effective engagement. This approach is particularly useful for developing scalable at-home CCT programs that do not require supervision by professional interventionists or family members, which often limit CCT accessibility or decrease outcomes from at-home CCT. Our combined clip-wise transformer and session-wise RNN classifier computer vision approach achieved the highest balanced accuracy, $F_1$ score, and precision compared to other state-of-the-art models for both detecting mental fatigue cases (binary classification) and estimating the level of mental fatigue (multi-class classification). The level of fatigue from this model was validated using the participant RT during CCT, which is well-established to increase with increasing mental fatigue. This measure provides a potential target for future closed-loop systems to manage mental fatigue during CCT and increase effective engagement. 

\paragraph{Comparisons to Image-only Static Convolutional Model.} As explained in the Results section, the previous state-of-the-art models \citep{zhao2020driver,guo2019pfld} employed image-only convolutional models, which encode frame-wise features and perform classification based on them. However, a crucial characteristic of emotional expressions is their nuanced unfolding over time, such as small movements of the muscles around the eye and some near the mouth. As a result, dynamic temporal information can play an important role in fatigue detection. Even for professionals who have undergone years of training in cognitive assessment, it is challenging to discern fatigue solely based on static images. To leverage dynamic temporal information, we utilize transformer architectures as shown in Table~\ref{tab:fatigue_results}. The results indicate that the transformer models generally outperform image-only static convolutional and our other ResNet baseline models.

\paragraph{Global Temporal Information.} As observed in Table~\ref{tab:fatigue_results}, the performance of RVT surpasses that of VT, with the key distinction that RVT incorporates an RNN layer to connect all unlabeled subsets of the cognitive training session (of the same person in the same session). This procedure better models the spectrum of engagement state on a global temporal scale throughout the entire session. Intuitively, individuals' facial appearances vary; hence, comparing the same subject across different time stages could reveal significant and personalized features, contributing to more accurate detection of fatigue. Our hierarchical structure of recurrent and transformer networks analyzes the local and global temporal information at different levels ensuring effective use of available ground truth while regulating the training procedure with the knowledge that domains of mental fatigue (e.g., attention, motivation, and affect) smoothly change over time.

\paragraph{Effective Engagement and Reaction Time.} One of the challenges of measuring effective engagement during CCT is a lack of continuous measures that can be collected without interfering with training. This is particularly true in older adults at risk for dementia who are often less tolerant of interruptions to training. A lack of continuous measures also makes validating novel measurements such as our facial expression fatigue composite score difficult. Reaction time is the most well-established objective measure of engagement with cognitive tasks such as CCT: reaction time is known to increase when participants are distracted or mind wandering~\citep{thomson2014link}. Reaction time also has a well-established relationship to mental fatigue: it shows a steady increase with time-on-task as mental fatigue increases and has been linked to self-report measures of mental fatigue in existing literature~\citep{jaydari2019mental}. The finding that our facial expression composite score (i.e., GIFS) shows a significant relationship to reaction time in a model including self-report mental fatigue is therefore a critical validation. This finding also has important clinical relevance: given that our composite shows a significant relationship to reaction time in a model including self-report mental fatigue, this suggests that our facial expression model explains additional variance in participant engagement over and above self-report. This means that not only does this method have the advantage of being able to continuously monitor fatigue but also can explain aspects of engagement related to mental fatigue that may be missed by self-report. Future research is needed to understand whether mental fatigue measured via facial expressions also relates to long-term gains from cognitive training, but the fact that our composite explained additional variance in reaction time compared to self-report suggests that this method has promise as a means of continuously monitoring a crucial aspect of effective engagement. 

\paragraph{Limitations.}
Although we introduced several novelties and contributions for fatigue detection and engagement adherence to cognitive training sessions, we acknowledge that there are several limitations: (1) \textit{Decreased fatigue detection performance with increased granularity:} As the granularity of fatigue ratings (i.e., the number of classes) increases, the performance of our model tends to decrease. For example, the RVT's performance on a 3-class classification task is significantly lower than that of a 2-class classification task. One reason for this could be the small sample size for classes with higher degrees of fatigue. This limitation may present challenges to the clinical utility of the models when a more granular fatigue rating is required. (2) \textit{Variability in individual performance:} The performance of our model varies across individual subjects. While it demonstrates acceptable performance for some test subjects, it falls short for some others who may not directly show signs of fatigue in their facial gestures. This highlights the need to improve the robustness of the mental fatigue model using supplemental data modalities (e.g., brain signals), as well as to better understand factors that contribute to individual differences. (3) \textit{Limited participant diversity:} Similarly to dementia research more broadly, our participants were limited in their socioeconomic and racial/ethnic diversity. To better validate the effectiveness of our model and to ensure future research aimed at developing personalized CCTs doesn't widen health disparities, it is necessary to recruit more diverse populations for future research. (4) \textit{Limited measures of mental fatigue}: while our model showed good performance in predicting mental fatigue and was validated using RT, pre- and post-session measures of mental fatigue do not provide the ideal training data. In the current study, it was critical not to interfere with CCT by measuring mental fatigue more frequently, however, future studies that collect mental fatigue more often during CCT (e.g., using experience sampling probes) may allow the development of a model with increased accuracy. (5) \textit{COVID-19 masking}: Participants wearing masks to reduce the risks from COVID-19 was essential during this study, however, this may also have led to reduced performance. 

\section{Conclusion}

Automatic real-time monitoring of mental fatigue lays the foundation for the future development of personalized closed-loop CCTs that adapt to manage mental fatigue and increase effective engagement to improve outcomes in older adults at risk for dementia. This is particularly important for developing scalable CCT interventions that do not require professional interventionist/family member supervision, making them more accessible to older adults in certain at-risk groups (e.g., those living alone). In this paper, we developed a combined clip-wise transformer and session-wise RNN classifier that achieved good performance (via balanced accuracy, $F_1$ score, and precision compared to other state-of-the-art models) and was validated by showing a significant relationship to the participant RT during CCT. This method lays the foundation for future work using real-time facial expression measures of mental fatigue during CCT in older adults with MCI to adapt training programs via closed-loop systems to manage mental fatigue, increase effective engagement, and improve outcomes from CCT. 

\section*{Competing Interests}
The authors declare no competing interests.

\section*{Data Availability}
The dataset has been approved IRB by the University of Rochester and Stanford University. Due to the sensitive nature of the data (including undeidentifiable facial images), the data is only available to the individuals listed on the IRB. 

\section*{Code Availability}
The code is available here: \url{https://github.com/PPWangyc/rvt}.

\section*{Acknowledgments}
This work was supported by several grants including NIH/NIA R61AG081723, NIH/NINR R01 NR015452, and Alzheimer’s Association AARG-22-926139. This research was also partially supported by the Stanford School of Medicine, Department of Psychiatry and Behavioral Sciences, Jaswa Innovator Award. Additionally, the authors would like to extend their gratitude to Ms. Mia Anthony for the data preparation of the parent study.

\bibliography{sample}

\begin{thebibliography}{34}
\providecommand{\natexlab}[1]{#1}
\providecommand{\url}[1]{\texttt{#1}}
\expandafter\ifx\csname urlstyle\endcsname\relax
  \providecommand{\doi}[1]{doi: #1}\else
  \providecommand{\doi}{doi: \begingroup \urlstyle{rm}\Url}\fi

\bibitem[National Academies~of Sciences et~al.(2017)National Academies~of Sciences, Medicine, et~al.]{national2017preventing}
Engineering National Academies~of Sciences, Medicine, et~al.
\newblock Preventing cognitive decline and dementia: A way forward.
\newblock 2017.

\bibitem[Turunen et~al.(2019)Turunen, Hokkanen, B{\"a}ckman, Stigsdotter-Neely, H{\"a}nninen, Paajanen, Soininen, Kivipelto, and Ngandu]{turunen2019computer}
Merita Turunen, Laura Hokkanen, Lars B{\"a}ckman, Anna Stigsdotter-Neely, Tuomo H{\"a}nninen, Teemu Paajanen, Hilkka Soininen, Miia Kivipelto, and Tiia Ngandu.
\newblock Computer-based cognitive training for older adults: Determinants of adherence.
\newblock \emph{PLoS One}, 14\penalty0 (7):\penalty0 e0219541, 2019.

\bibitem[Lampit et~al.(2014)Lampit, Hallock, and Valenzuela]{lampit2014computerized}
Amit Lampit, Harry Hallock, and Michael Valenzuela.
\newblock Computerized cognitive training in cognitively healthy older adults: a systematic review and meta-analysis of effect modifiers.
\newblock \emph{PLoS medicine}, 11\penalty0 (11):\penalty0 e1001756, 2014.

\bibitem[Yardley et~al.(2016)Yardley, Spring, Riper, Morrison, Crane, Curtis, Merchant, Naughton, and Blandford]{yardley2016understanding}
Lucy Yardley, Bonnie~J Spring, Heleen Riper, Leanne~G Morrison, David~H Crane, Kristina Curtis, Gina~C Merchant, Felix Naughton, and Ann Blandford.
\newblock Understanding and promoting effective engagement with digital behavior change interventions.
\newblock \emph{American journal of preventive medicine}, 51\penalty0 (5):\penalty0 833--842, 2016.

\bibitem[Stern(2012)]{stern2012cognitive}
Yaakov Stern.
\newblock Cognitive reserve in ageing and alzheimer's disease.
\newblock \emph{The Lancet Neurology}, 11\penalty0 (11):\penalty0 1006--1012, 2012.

\bibitem[Lin(2022)]{lin2022multi}
Feng~V Lin.
\newblock A multi-dimensional model of fatigue in old age: Implications for brain aging, 2022.

\bibitem[Kukla et~al.(2022)Kukla, Anthony, Chen, Turnbull, Baran, and Lin]{kukla2022brain}
Bennett Kukla, Mia Anthony, Shuyi Chen, Adam Turnbull, Timothy~M Baran, and Feng~V Lin.
\newblock Brain small-worldness properties and perceived fatigue in mild cognitive impairment.
\newblock \emph{The Journals of Gerontology: Series A}, 77\penalty0 (3):\penalty0 541--546, 2022.

\bibitem[Huang et~al.(2023)Huang, Roth, Cidav, Chung, Amjad, Thorpe~Jr, Boyd, and Cudjoe]{huang2023social}
Alison~R Huang, David~L Roth, Tom Cidav, Shang-En Chung, Halima Amjad, Roland~J Thorpe~Jr, Cynthia~M Boyd, and Thomas~KM Cudjoe.
\newblock Social isolation and 9-year dementia risk in community-dwelling medicare beneficiaries in the united states.
\newblock \emph{Journal of the American Geriatrics Society}, 71\penalty0 (3):\penalty0 765--773, 2023.

\bibitem[Hill et~al.(2015)Hill, P{\'e}rez-Stable, Anderson, and Bernard]{hill2015national}
Carl~V Hill, Eliseo~J P{\'e}rez-Stable, Norman~A Anderson, and Marie~A Bernard.
\newblock The national institute on aging health disparities research framework.
\newblock \emph{Ethnicity \& disease}, 25\penalty0 (3):\penalty0 245, 2015.

\bibitem[Mlynski et~al.(2021)Mlynski, Reza, Whitted, Cox, Garsea, and Wright]{mlynski2021fatigue}
Christopher Mlynski, Ariel Reza, Melissa Whitted, Caytlin Cox, Anne Garsea, and Rex~A Wright.
\newblock Fatigue influence on inhibitory control: Cardiovascular and performance findings elucidate the role of restraint intensity.
\newblock \emph{Psychophysiology}, 58\penalty0 (9):\penalty0 e13881, 2021.

\bibitem[Kong et~al.(2021)Kong, Posada-Quintero, Daley, Chon, and Bolkhovsky]{kong2021facial}
Youngsun Kong, Hugo~F Posada-Quintero, Matthew~S Daley, Ki~H Chon, and Jeffrey Bolkhovsky.
\newblock Facial features and head movements obtained with a webcam correlate with performance deterioration during prolonged wakefulness.
\newblock \emph{Attention, Perception, \& Psychophysics}, 83:\penalty0 525--540, 2021.

\bibitem[Gu et~al.(2002)Gu, Ji, and Zhu]{gu2002active}
Haisong Gu, Qiang Ji, and Zhiwei Zhu.
\newblock Active facial tracking for fatigue detection.
\newblock In \emph{Sixth IEEE Workshop on Applications of Computer Vision, 2002.(WACV 2002). Proceedings.}, pages 137--142. IEEE, 2002.

\bibitem[Li et~al.(2019)Li, Wang, Du, Huang, Feng, and Zhou]{li2019accurate}
Kangning Li, Shangshang Wang, Chang Du, Yuxin Huang, Xin Feng, and Fengfeng Zhou.
\newblock Accurate fatigue detection based on multiple facial morphological features.
\newblock \emph{Journal of Sensors}, 2019, 2019.

\bibitem[Dosovitskiy et~al.(2021)Dosovitskiy, Beyer, Kolesnikov, Weissenborn, Zhai, Unterthiner, Dehghani, Minderer, Heigold, Gelly, Uszkoreit, and Houlsby]{dosovitskiy2021an}
Alexey Dosovitskiy, Lucas Beyer, Alexander Kolesnikov, Dirk Weissenborn, Xiaohua Zhai, Thomas Unterthiner, Mostafa Dehghani, Matthias Minderer, Georg Heigold, Sylvain Gelly, Jakob Uszkoreit, and Neil Houlsby.
\newblock An image is worth 16x16 words: Transformers for image recognition at scale.
\newblock In \emph{International Conference on Learning Representations}, 2021.
\newblock URL \url{https://openreview.net/forum?id=YicbFdNTTy}.

\bibitem[Sherstinsky(2020)]{sherstinsky2020fundamentals}
Alex Sherstinsky.
\newblock Fundamentals of recurrent neural network (rnn) and long short-term memory (lstm) network.
\newblock \emph{Physica D: Nonlinear Phenomena}, 404:\penalty0 132306, 2020.

\bibitem[Jaydari~Fard et~al.(2019)Jaydari~Fard, Tahmasebi~Boroujeni, and Lavender]{jaydari2019mental}
Saeed Jaydari~Fard, Shahzad Tahmasebi~Boroujeni, and Andrew~P Lavender.
\newblock Mental fatigue impairs simple reaction time in non-athletes more than athletes.
\newblock \emph{Fatigue: Biomedicine, Health \& Behavior}, 7\penalty0 (3):\penalty0 117--126, 2019.

\bibitem[Lin et~al.(2021)Lin, Heffner, Gevirtz, Zhang, Tadin, and Porsteinsson]{lin2021targeting}
Feng~V Lin, Kathi Heffner, Richard Gevirtz, Zhengwu Zhang, Duje Tadin, and Anton Porsteinsson.
\newblock Targeting autonomic flexibility to enhance cognitive training outcomes in older adults with mild cognitive impairment: study protocol for a randomized controlled trial.
\newblock \emph{Trials}, 22\penalty0 (1):\penalty0 1--15, 2021.

\bibitem[Nasreddine et~al.(2005)Nasreddine, Phillips, B{\'e}dirian, Charbonneau, Whitehead, Collin, Cummings, and Chertkow]{nasreddine2005montreal}
Ziad~S Nasreddine, Natalie~A Phillips, Val{\'e}rie B{\'e}dirian, Simon Charbonneau, Victor Whitehead, Isabelle Collin, Jeffrey~L Cummings, and Howard Chertkow.
\newblock The montreal cognitive assessment, moca: a brief screening tool for mild cognitive impairment.
\newblock \emph{Journal of the American Geriatrics Society}, 53\penalty0 (4):\penalty0 695--699, 2005.

\bibitem[Jiang et~al.(2020)Jiang, Zong, Zheng, Tang, Xia, Lu, and Liu]{10.1145/3394171.3413620}
Xingxun Jiang, Yuan Zong, Wenming Zheng, Chuangao Tang, Wanchuang Xia, Cheng Lu, and Jiateng Liu.
\newblock Dfew: A large-scale database for recognizing dynamic facial expressions in the wild.
\newblock In \emph{Proceedings of the 28th ACM International Conference on Multimedia}, MM '20, page 2881–2889, New York, NY, USA, 2020. Association for Computing Machinery.
\newblock ISBN 9781450379885.
\newblock \doi{10.1145/3394171.3413620}.
\newblock URL \url{https://doi.org/10.1145/3394171.3413620}.

\bibitem[Lee et~al.(1991)Lee, Hicks, and Nino-Murcia]{lee1991validity}
Kathryn~A Lee, Gregory Hicks, and German Nino-Murcia.
\newblock Validity and reliability of a scale to assess fatigue.
\newblock \emph{Psychiatry research}, 36\penalty0 (3):\penalty0 291--298, 1991.

\bibitem[Lin et~al.(2016)Lin, Ren, Cotton, Porsteinsson, Mapstone, and Heffner]{lin2016mental}
Feng Lin, Ping Ren, Kelly Cotton, Anton Porsteinsson, Mark Mapstone, and Kathi~L Heffner.
\newblock Mental fatigability and heart rate variability in mild cognitive impairment.
\newblock \emph{The American Journal of Geriatric Psychiatry}, 24\penalty0 (5):\penalty0 374--378, 2016.

\bibitem[Masel et~al.(2007)Masel, King, and Maughan]{masel2007loss}
Joanna Masel, Oliver~D King, and Heather Maughan.
\newblock The loss of adaptive plasticity during long periods of environmental stasis.
\newblock \emph{The American Naturalist}, 169\penalty0 (1):\penalty0 38--46, 2007.

\bibitem[He et~al.(2016)He, Zhang, Ren, and Sun]{he2016deep}
Kaiming He, Xiangyu Zhang, Shaoqing Ren, and Jian Sun.
\newblock Deep residual learning for image recognition.
\newblock In \emph{Proceedings of the IEEE conference on computer vision and pattern recognition}, pages 770--778, 2016.

\bibitem[Zhao and Liu(2021)]{zhao2021former}
Zengqun Zhao and Qingshan Liu.
\newblock Former-dfer: Dynamic facial expression recognition transformer.
\newblock In \emph{Proceedings of the 29th ACM International Conference on Multimedia}, pages 1553--1561, 2021.

\bibitem[Chen et~al.(2020)Chen, Yang, Rooks, Anthony, Zhang, Tadin, Heffner, and Lin]{chen2020autonomic}
Quanjing Chen, Haichuan Yang, Brian Rooks, Mia Anthony, Zhengwu Zhang, Duje Tadin, Kathi~L Heffner, and Feng~V Lin.
\newblock Autonomic flexibility reflects learning and associated neuroplasticity in old age.
\newblock \emph{Human brain mapping}, 41\penalty0 (13):\penalty0 3608--3619, 2020.

\bibitem[Zhao et~al.(2020)Zhao, Zhou, Zhang, Yan, Xu, Zhang, et~al.]{zhao2020driver}
Zuopeng Zhao, Nana Zhou, Lan Zhang, Hualin Yan, Yi~Xu, Zhongxin Zhang, et~al.
\newblock Driver fatigue detection based on convolutional neural networks using em-cnn.
\newblock \emph{Computational intelligence and neuroscience}, 2020, 2020.

\bibitem[Guo et~al.(2019)Guo, Li, Yu, Zhang, Ma, Ma, Liu, and Ling]{guo2019pfld}
Xiaojie Guo, Siyuan Li, Jinke Yu, Jiawan Zhang, Jiayi Ma, Lin Ma, Wei Liu, and Haibin Ling.
\newblock Pfld: A practical facial landmark detector.
\newblock \emph{arXiv preprint arXiv:1902.10859}, 2019.

\bibitem[Pham et~al.(2021)Pham, Vu, and Tran]{pham2021facial}
Luan Pham, The~Huynh Vu, and Tuan~Anh Tran.
\newblock Facial expression recognition using residual masking network.
\newblock In \emph{2020 25Th international conference on pattern recognition (ICPR)}, pages 4513--4519. IEEE, 2021.

\bibitem[Goodfellow et~al.(2013)Goodfellow, Erhan, Carrier, Courville, Mirza, Hamner, Cukierski, Tang, Thaler, Lee, et~al.]{goodfellow2013challenges}
Ian~J Goodfellow, Dumitru Erhan, Pierre~Luc Carrier, Aaron Courville, Mehdi Mirza, Ben Hamner, Will Cukierski, Yichuan Tang, David Thaler, Dong-Hyun Lee, et~al.
\newblock Challenges in representation learning: A report on three machine learning contests.
\newblock In \emph{Neural Information Processing: 20th International Conference, ICONIP 2013, Daegu, Korea, November 3-7, 2013. Proceedings, Part III 20}, pages 117--124. Springer, 2013.

\bibitem[Wilcoxon(1992)]{wilcoxon1992individual}
Frank Wilcoxon.
\newblock Individual comparisons by ranking methods.
\newblock In \emph{Breakthroughs in statistics}, pages 196--202. Springer, 1992.

\bibitem[Wang et~al.(2021)Wang, Liu, Hu, Shi, and Mei]{wang2021facex}
Jun Wang, Yinglu Liu, Yibo Hu, Hailin Shi, and Tao Mei.
\newblock Facex-zoo: A pytorh toolbox for face recognition.
\newblock 2021.

\bibitem[Thomson et~al.(2014)Thomson, Seli, Besner, and Smilek]{thomson2014link}
David~R Thomson, Paul Seli, Derek Besner, and Daniel Smilek.
\newblock On the link between mind wandering and task performance over time.
\newblock \emph{Consciousness and cognition}, 27:\penalty0 14--26, 2014.

\bibitem[Deb(2022)]{Deb_Face_Mask_Detection_2022}
Chandrika Deb.
\newblock {Face Mask Detection}, 2 2022.
\newblock URL \url{https://github.com/chandrikadeb7/Face-Mask-Detection}.

\bibitem[Vaswani et~al.(2017)Vaswani, Shazeer, Parmar, Uszkoreit, Jones, Gomez, Kaiser, and Polosukhin]{vaswani2017attention}
Ashish Vaswani, Noam Shazeer, Niki Parmar, Jakob Uszkoreit, Llion Jones, Aidan~N Gomez, {\L}ukasz Kaiser, and Illia Polosukhin.
\newblock Attention is all you need.
\newblock \emph{Advances in neural information processing systems}, 30, 2017.

\end{thebibliography}
\bibliographystyle{unsrtnat} 

\newpage
\appendix
%
\label{sec:appendix}
\renewcommand{\thefigure}{S\arabic{figure}}
\renewcommand{\thetable}{S\arabic{table}}  
    \makeatletter 
       \setcounter{table}{0}
       \renewcommand{\thetable}{S\@arabic\c@table}
       \setcounter{figure}{0}
       \renewcommand{\thefigure}{S\@arabic\c@figure}
    \makeatother
%

\section{Methodological Details}
\begin{figure}[t]
    \centering
    \includegraphics[width=1\linewidth]{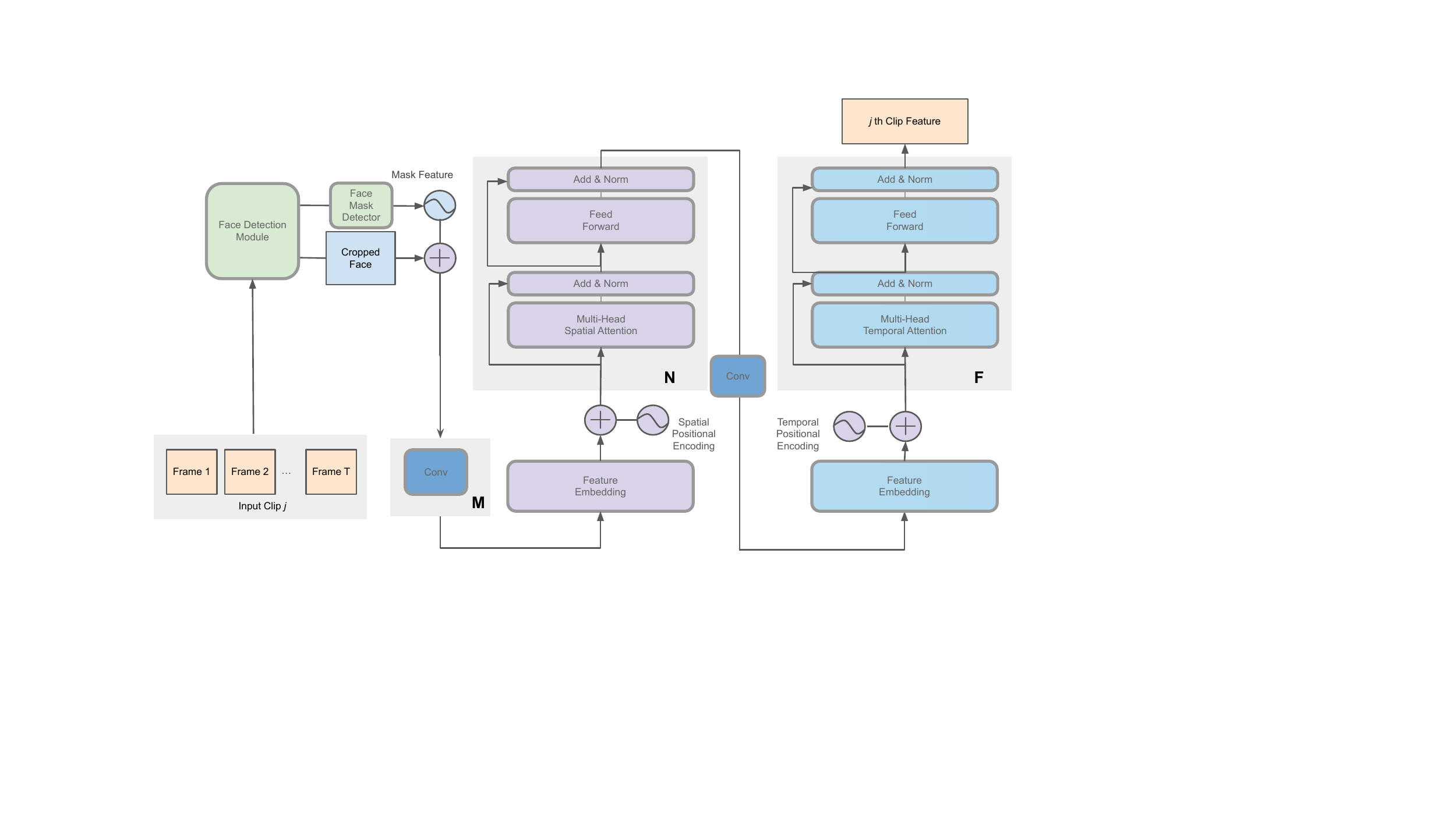}\vspace{-15pt}
    \caption{Details of our Transformer Encoder. It includes a mask detection module, $M$ blocks of convolution, $N$ spatial attention modules, and $F$ temporal attention components. The transformer encoder outputs a feature vector for each clip.}
    \label{fig:tvc}
\end{figure}

As shown in Figure~\ref{fig:tvc}, our transformer encoder component consists of three modules, a face detection module with a face mask detector, a spatial transformer encoder, and a temporal transformer encoder.  

For any $x$ in $X \in \mathcal{X}$, we use the face detection module to get the cropped face image $x^f \in \mathcal{R}^{112 \times 111}$ and the face mask feature $x^m \in \{0,1\}$, where $1$ ($0$)  indicates the subject is (not) wearing a face mask in the frame $x$. The face detection module is based on ResNet50~\citep{he2016deep} and is pre-trained by \cite{Deb_Face_Mask_Detection_2022} on their self-collected dataset. The output of the face detection module is $x' \in \mathcal{R}^{112 \times 112}$, which equals to $ [x^f \; {x^m}^{112 \times 1}]$. If we put all the $x'$ for all the $x$ in $X$ together, we are able to obtain $X' \in \mathcal{R}^{T\times 112 \times 112}$, which is the input to the transformer module. 

The spatial transformer encoder and temporal transformer encoder are pre-trained by \cite{zhao2021former} on a dataset for Recognizing Dynamic Facial Expressions in the Wild (DFEW)~\citep{10.1145/3394171.3413620}. DFEW is a dataset of video clips containing facial expressions of seven basic emotions, created for evaluating facial expression recognition algorithms. It contains over 1,000 annotated video clips recorded under controlled conditions with diverse actors. They used a combination of convolutional and self-attentional layers to extract spatial and temporal features from facial images and videos. The network, called Former-DFER, is designed to dynamically recognize seven basic emotions (anger, disgust, fear, happiness, sadness, surprise, and neutral) with high accuracy in real-time applications. 

There are $N$ encoders that serve as the spatial transformer encoder. Given any $X'$ from the face detection module, the first encoder extracts feature maps via $M$ convolution blocks (Here $M=4$), and then flattens it to $M^f \in \mathcal{R}^{Q \times C} $,  where $Q$ is the number of word embedding and $C$ is the length of each embedding. Therefore, the input embedding to the spatial transformer is $z_p^0=m^f_p+e_p$, where $p \in{1,2,\cdots,Q}$, and $e_p \in \mathcal{R}^C$. Then, as stated in \citep{10.1145/3394171.3413620,vaswani2017attention}, Query, Key, and Value could be computed and passed to the self-attention computation, then we can get the output $X^S \in \mathcal{R}^S$ at the end of the encoder. In a similar way, the $X^S$ was given to the temporal encoder after passing convolutional layer(s), and the encoded feature $X^E$ is the clip feature ($i^\text{th}$ clip feature in the Figure~\ref{fig:tvc}) obtained from the pre-trained Transformer Encoder.



\end{document}